\documentclass[journal, letterpaper]{IEEEtran}

\usepackage{graphicx}
\usepackage{url}        
\usepackage{amsmath}    
\usepackage{textgreek}
\usepackage{listings}
\usepackage{csvsimple}
\usepackage{longtable}

\usepackage[table]{xcolor}
\usepackage[table,dvipsnames]{xcolor} 
\usepackage[most]{tcolorbox}
\usepackage{tabularx} 
\usepackage{colortbl}     
\definecolor{lightgray}{gray}{0.96} 
\usepackage[normalem]{ulem} 
\usepackage{booktabs}
\usepackage[ruled,vlined,linesnumbered]{algorithm2e}
\usepackage{adjustbox}
\usepackage[table]{xcolor}
\usepackage{multirow}
\usepackage{pifont}
\definecolor{darkred}{RGB}{220,20,60}
\usepackage{amssymb}

\usepackage[title]{appendix}
\usepackage{hyperref}

\usepackage[utf8]{inputenc}
\usepackage{authblk}
\usepackage{bbding}

\begin{document}

\title{FashionPose: Unified Text-Driven Fashion Synthesis \\with Joint Geometric and Photometric Control}

\author[1]{Chuancheng Shi}
\author[2]{Shaotian Li}
\author[3]{Zhenlong Yuan}
\author[4]{Zifeng Cheng}
\author[5]{Fei Shen\textsuperscript{\Envelope}}

\affil[1]{The University of Sydney}
\affil[2]{Macquarie University}
\affil[3]{Chinese Academy of Sciences}
\affil[4]{Nanjing University}
\affil[5]{NExT++ Research Centre, National University of Singapore}

\affil[ ]{\dag ~ Equal Contribution}
\affil[ ]{\Envelope ~ Corresponding Author}

\maketitle

\begin{abstract}
Realistic and controllable garment synthesis is essential for fashion e-commerce, yet it demands precise coordination between human pose geometry and environmental photometry. 
Conventional pose-guided frameworks suffer from two fundamental limitations: they rely heavily on predefined skeletons from off-the-shelf estimators, restricting semantic flexibility; and they predominantly focus on "studio-like" generation under neutral lighting, failing to reconcile geometric configurations with complex, scene-specific illumination described in natural language. 
To bridge this gap, we propose \textbf{FashionPose}, a cascaded architecture that reconciles geometric and photometric control within a unified language-driven interface. 
Unlike conventional frameworks, our framework employs a decoupled yet synergistic strategy: 
(1) a bidirectional contrastive alignment mechanism that grounds textual semantics into an explicit geometric manifold, enabling template-free pose generation; 
(2) an identity-anchored synthesis module that translates these geometric priors into high-fidelity imagery while preserving fine-grained appearance; 
and (3) a prompt-conditioned relighting module that leverages the generated pose as a spatial anchor to achieve environment-aware shading. 
This hierarchical design effectively transforms high-level instructions into consistent visual representations, ensuring both structural precision and atmospheric harmony.
To facilitate this paradigm, we construct PoseCap, a dataset with over 40k caption–keypoint pairs. Extensive experiments demonstrate that FashionPose outperforms existing benchmarks in pose accuracy and physical realism, providing a robust solution for personalized, scene-aware virtual fashion displays.
\end{abstract}




\section{Introduction}

\begin{figure}[t]
    \centering
    \includegraphics[width=1\linewidth]{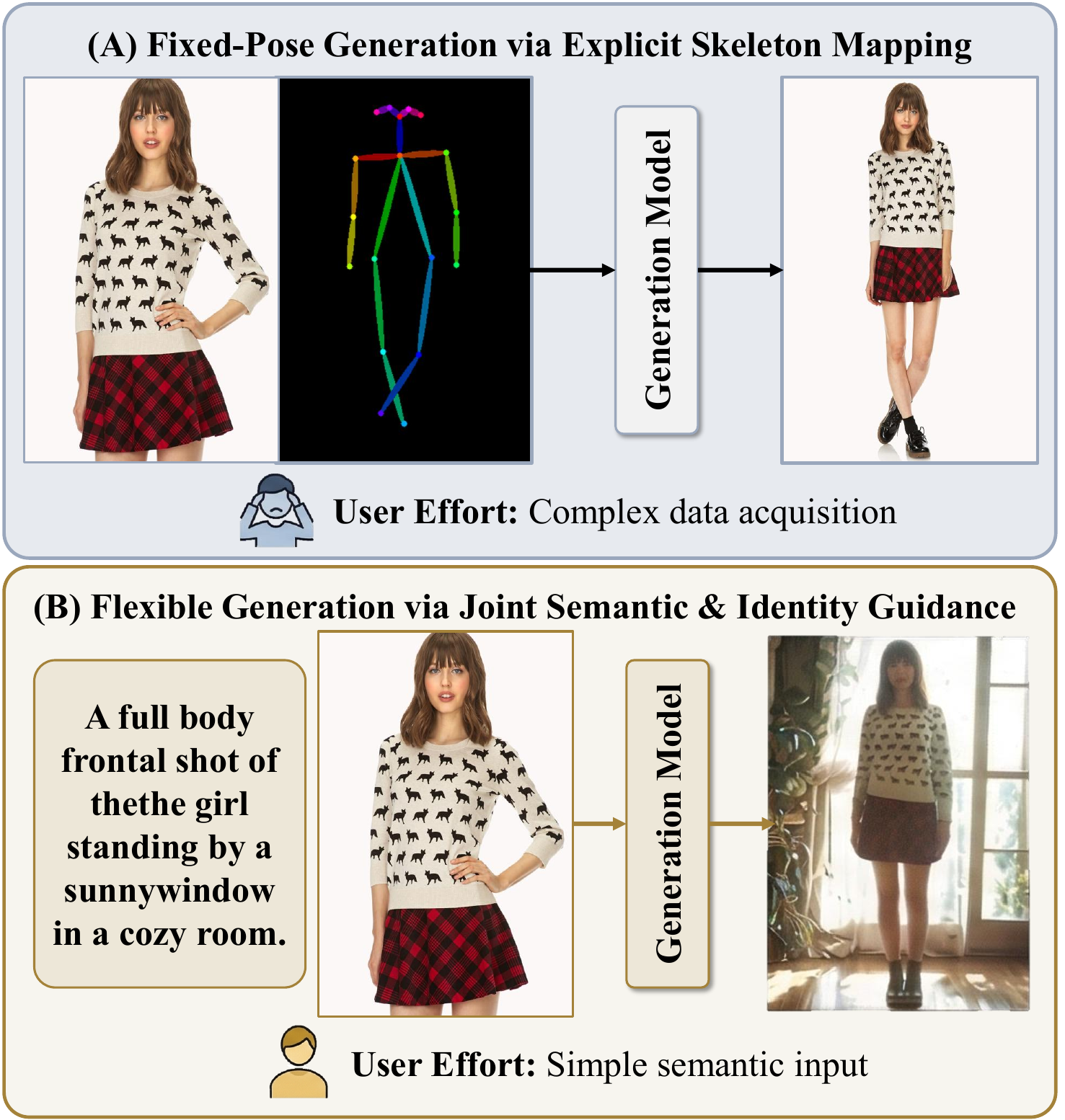}
    \caption{Comparison between existing pose-guided frameworks and our approach. While conventional methods require additional preprocessing to obtain an explicit pose skeleton, our method uses a single prompt to specify both pose and lighting.}
    \label{fig:intro}
    \vspace{-0.3cm}
\end{figure}

In the domain of realistic fashion e-commerce, the ideal interaction paradigm is one where users can dictate pose and lighting outcomes through natural language alone. 
This approach allows users to seamlessly specify body pose, action intent, and environmental lighting. Such a language-first interaction offers superior scalability compared to manual annotation or retrieval-based methods. However, a fundamental disconnect persists in current methodologies. Existing pose-guided generation frameworks~\cite{shen2023advancing,shen2024imagpose,zhang2022exploring,bhunia2023person,lu2024coarse} predominantly rely on explicit, pre-defined pose templates or external skeleton estimators during the inference stage. This rigid dependency limits semantic flexibility and fails to align geometric configurations with the complex, scene-specific photometry described in the text. Consequently, there is an imperative need for a unified framework that establishes language as the primary interface for governing both structural geometry and photometric conditions, thereby eliminating the reliance on external priors.

As illustrated in Figure~\ref{fig:intro}, conventional pose-guided frameworks strictly rely on explicit skeletons as a prerequisite, which are typically derived from off-the-shelf pose estimators. Beyond the overhead of preprocessing, a more critical limitation lies in the absence of environment-aware control. Current frameworks operate under a "studio-like" assumption. They prioritize geometric alignment while defaulting to neutral or implicit lighting conditions. These systems lack a principled mechanism to jointly interpret body pose and the diverse illumination contexts described in language, such as specific shadow directions or outdoor ambience. Consequently, when a prompt dictates both a specific pose and an environmental context, these systems fail to maintain the necessary coupling between geometry and photometry. This failure results in synthetic characters that appear artificially superimposed onto backgrounds with inconsistent shading. Such visual dissonance stems from the disparity between the holistic intent of the language and framework-level controls that remain constrained to isolated geometric conditions.

To bridge this gap, we propose FashionPose, the first unified framework that redefines natural language as the primary interface for joint geometric and photometric control. Departing from existing studio-centric paradigms, our approach ensures environmental consistency through a cascaded three-stage framework. First, we implement a language-driven pose generator that regresses explicit 2D skeletons via bidirectional contrastive alignment. This design eliminates the reliance on external templates. Second, we employ a multi-scale pose-conditioned diffusion model to synthesize high-fidelity images. This model uses token-based appearance guidance to effectively anchor identity, even under drastic pose changes. Third, to ensure realistic integration beyond studio settings, we introduce a prompt-conditioned relighting module. By leveraging an identity-anchor strategy, this module harmonizes the synthesized character with the complex environmental lighting described in the text. To support this data-demanding task, we also construct PoseCap, a large-scale dataset comprising over 40k caption-keypoint pairs. 

In summary, our work makes four primary contributions:

\begin{itemize}
  \item We propose FashionPose, the first unified framework that establishes natural language as the comprehensive interface for joint geometric and photometric control. This effectively bridges the gap between rigid pose inputs and environment-aware synthesis.
    
  \item We introduce a novel bidirectional contrastive alignment mechanism. This enables the direct regression of explicit 2D skeletons from semantic descriptions without relying on external pose templates.
    
  \item We design a prompt-conditioned relighting module utilizing an identity-anchor strategy. This component harmonizes the synthesized character with complex, scene-specific lighting conditions while rigorously preserving garment fidelity.
    
  \item We present PoseCap, a large-scale dataset comprising over 40k high-quality caption-keypoint pairs, providing a robust benchmark to facilitate future research in language-guided human generation.
\end{itemize}


\section{Related Work}

\noindent\textbf{Pose-Guided Person Image Synthesis.}
Pose-guided person image synthesis focuses on generating realistic human images conditioned on keypoints or pose maps. Since the emergence of GAN-based frameworks \cite{ma2017pose}, research has evolved to enhance spatial alignment, identity preservation, and garment detail via sophisticated feature fusion and latent-space manipulation \cite{harkonen2020ganspace}. Recent advancements further push the boundaries of high-frequency structure and long-range consistency for fashion-oriented applications \cite{shen2025long}. However, practical deployment remains constrained by two major hurdles: the lack of accessible, low-expertise pose acquisition mechanisms, as users currently rely on cumbersome manual specifications or templates, and the absence of illumination-aware refinement, which is essential for ensuring photorealistic harmony between the subject and the environment \cite{zhang2025scaling}. To address this critical gap, emerging research is prioritizing advanced relighting mechanisms over traditional static appearance transfer~\cite{wang2025comprehensive}. Proper illumination-aware synthesis is increasingly recognized not merely as an optional post-processing step, but as a core component of the generation pipeline that dictates the perceived realism~\cite{ren2026sceneshine,chen2025polar}. Consequently, contemporary methods are actively exploring ways to disentangle lighting representations from texture and identity, allowing for precise, dynamic control over environmental shading and shadows~\cite{chaturvedi2025synthlight}. Ultimately, mastering this relighting capability is pivotal for seamlessly integrating synthesized human figures into diverse, real-world scenes without jarring visual artifacts.

\noindent\textbf{Text to Human Pose Image.}
To simplify pose specification, natural language has emerged as a user-friendly interface, bridging the gap between everyday expressions and complex body configurations. Recent research uses transformer- and diffusion-based text-conditioned models to generate or edit pose or motion sequences from free-form descriptions \cite{liao2025shapemymoves,uchida2025mola,li2025simmotionedit}, while adversarial learning and retrieval augmentation are explored to strengthen semantic grounding and reduce text–motion mismatch \cite{uchida2025mola,kalakonda2025morag,li2025remomask}. However, adapting these generated skeletons for high-fidelity image synthesis remains challenging because text-only inputs often under-specify fine-grained local geometry, so practical systems introduce explicit geometric anchors such as SMPL or pose maps and extra supervision or domain adaptation to preserve visual fidelity \cite{buchheim2025controlling,jin2025textdrivenhumanimage}. Moreover, many frameworks\cite{jin2025textdrivenhumanimage,choi2025finecontrolnet} are staged as text-to-pose followed by pose-conditioned synthesis, so upstream pose errors can propagate to the final image. Finally, pose control alone does not determine instance-level appearance, motivating additional appearance or identity-focused conditioning for realistic and identity-consistent generation.

\begin{figure}[t]
  \centering
  \includegraphics[width=1\linewidth]{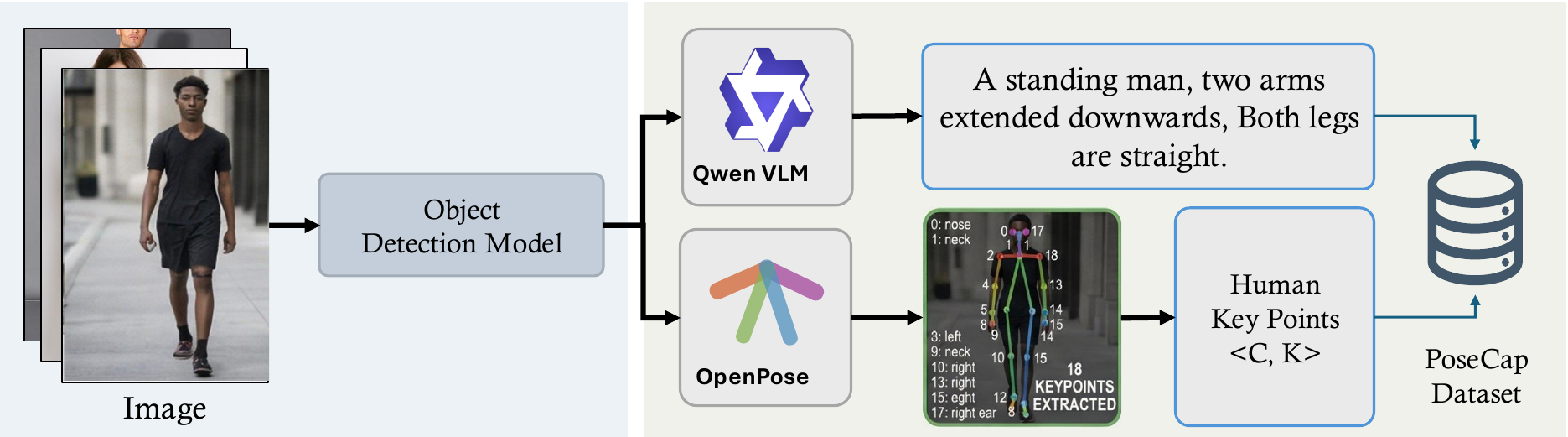}
  \vspace{-0.3cm}
  \caption{\textbf{Data curation framework for PoseCap.} The framework processes raw images through human detection and resizing, followed by parallel annotation using Qwen VLM for captions and OpenPose for keypoints to generate semantic-structural pairs.}
  \label{fig:data}
  \vspace{-0.2cm}
\end{figure}

\begin{figure}[t]
    \centering
    \includegraphics[width=1\linewidth]{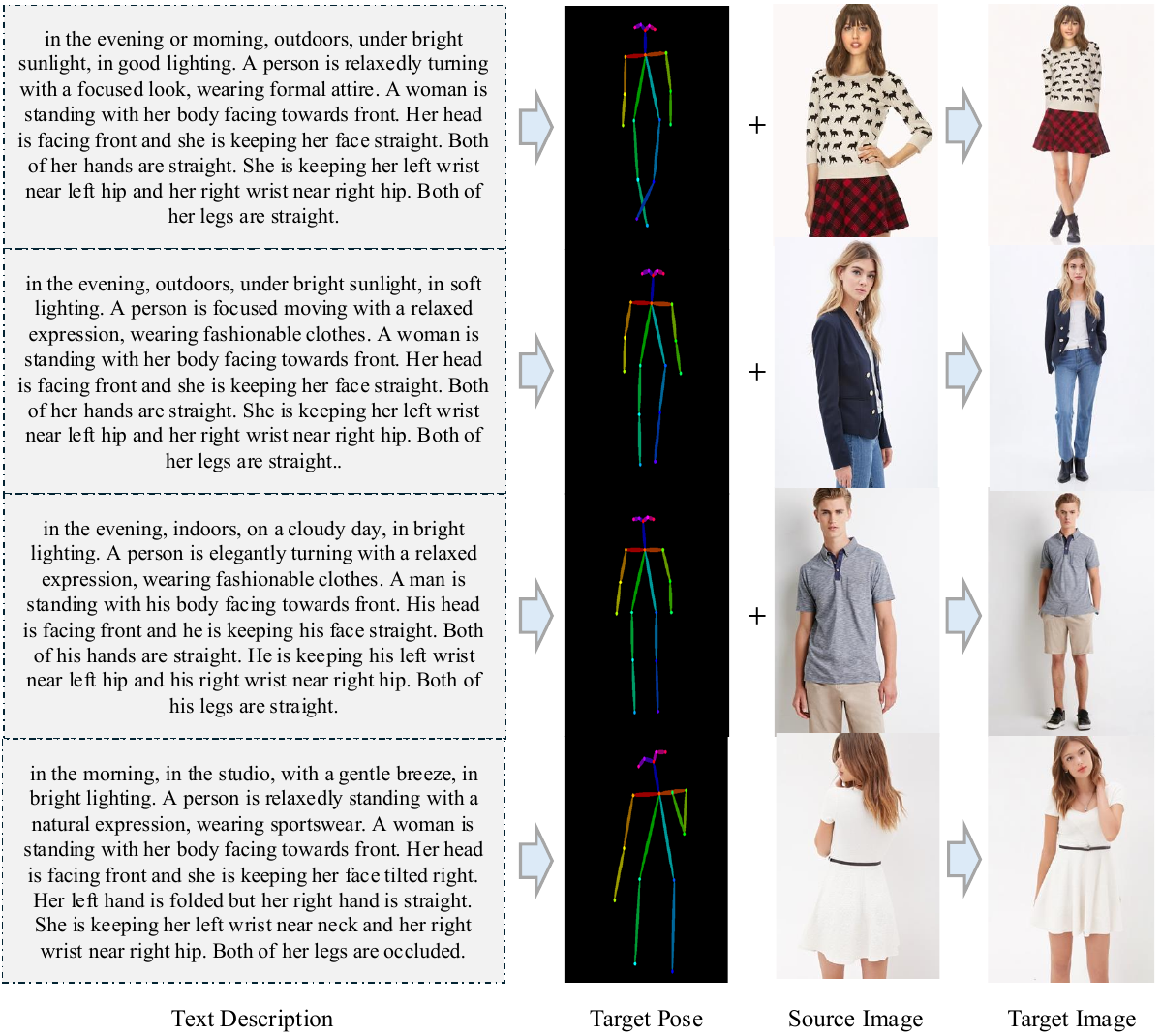}
    \caption{\textbf{Illustration of the PoseCap dataset structure.} Each data sample consists of a high-quality image-pose-text triplet. }
    \label{fig:dataset_st}
    \vspace{-0.3cm}
\end{figure}

\begin{figure*}[t]
  \centering
  \includegraphics[width=1\linewidth]{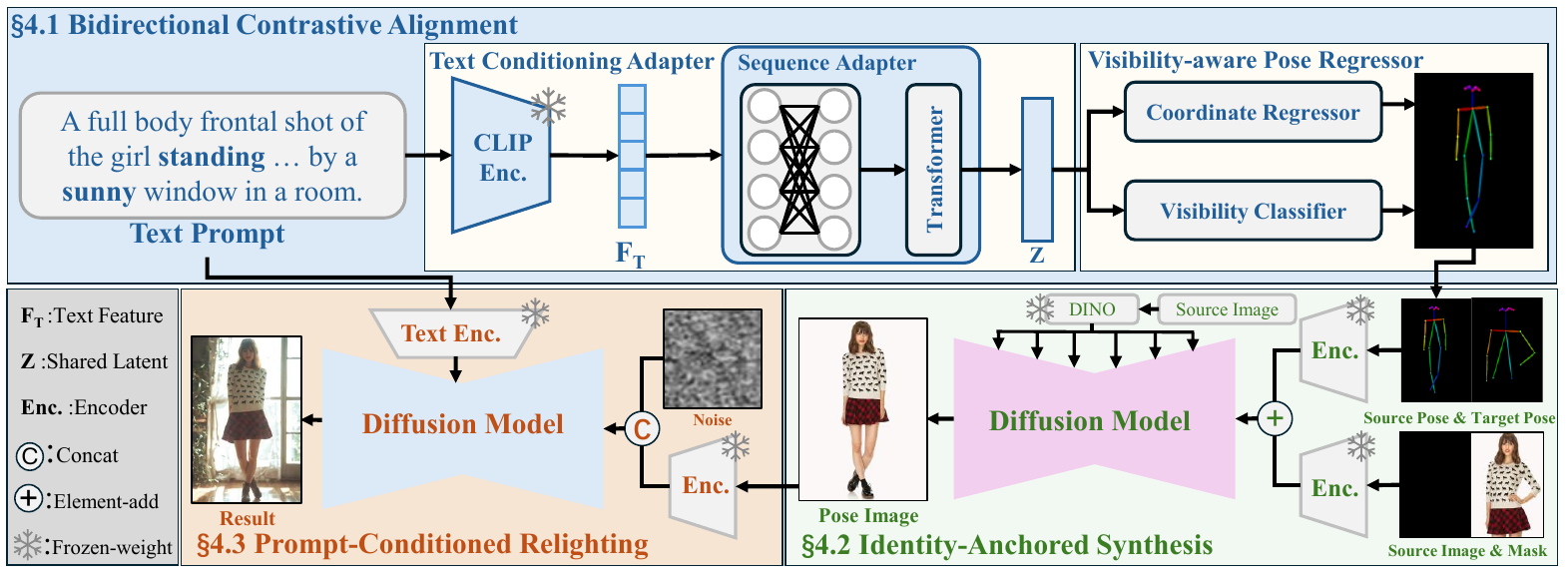}
  \caption{
    \textbf{The overall architecture of the proposed framework.} The framework operates in three stages: (a) bidirectional contrastive alignment maps a text prompt to a target pose using a Text Conditioning Adapter to encode features and a visibility-aware pose regressor to predict coordinates and visibility; (b) identity-anchored synthesizes a pose-aligned image conditioned on the predicted pose and a source image via a diffusion model; and (c) Prompt-Conditioned Relighting adjusts the illumination of the generated result according to the text prompt to produce the final output.
  }
  \label{fig:model_architecture}
\end{figure*}

\section{PoseCap Dataset}
To address the scarcity of high-quality text-pose paired data, we constructed PoseCap (see Figure~\ref{fig:data}), a specialized corpus comprising over 40k aligned image-caption-keypoint pairs derived from DeepFashion~\cite{liu2016deepfashion}. The construction framework adheres to a rigorous quality control protocol. First, we filter raw samples using YOLOX~\cite{ge2021yolox} to identify single-person subjects with clear visibility, discarding cluttered or multi-person scenes, and resize valid images to a uniform $512\times512$ resolution. To bridge the gap between abstract semantics and physical geometry, we generate rich multimodal supervision by prompting the Qwen VLM~\cite{bai2023qwenvlversatilevisionlanguagemodel} to produce descriptive captions ($C$) that explicitly detail limb orientations and body gestures. Simultaneously, we extract 18 precise anatomical landmarks ($K$) via OpenPose~\cite{cao2019openpose}. The final dataset is stored as lightweight $\langle C, K \rangle$ tuples, effectively decoupling geometric structure from pixel appearance to facilitate focused, structure-aware training.

\noindent\textbf{Details of PoseCap Dataset.}
As shown in Figure~\ref{fig:dataset_st}, to facilitate the training of FashionPose, we constructed PoseCap, a large-scale dataset featuring aligned image-pose-text triplets. The dataset is organized in a standard JSON format, where each entry $D_i = \{I_i, P_i, T_i\}$ encapsulates the essential components for unified synthesis. Specifically, each data sample contains the relative path to the preprocessed high-resolution image alongside a flattened array of 18 human joint coordinates extracted via OpenPose, which serves as the ground truth for our geometric alignment module. Crucially, unlike standard datasets that utilize generic descriptions, our semantic annotations are densely structured to bridge abstract semantics with physical constraints. As shown in Figure~\ref{fig:dataset_st} of the main paper, these captions are hierarchically structured to explicitly encode photometric context (e.g., lighting conditions and scene environment), subject attributes (e.g., clothing style and expression), and fine-grained geometry (e.g., detailed limb orientations and body orientation). This structured textual representation enables the model to decouple and effectively control both environmental illumination and geometric configuration within a unified framework.

\begin{table*}[t]
\centering
\caption{\textbf{Quantitative comparison of the end-to-end framework against SOTA baselines and a modern foundation model (IC-Light) on PoseCap dataset.} \textbf{Aes-S} denotes the VLM-based Aesthetic Score (higher is better), reflecting the perceptual quality. }
\label{tab:final_comparison}
\resizebox{\textwidth}{!}{
    \begin{tabular}{l|cc|ccccc}
    \hline
    \multicolumn{1}{c|}{\textbf{Method Configuration}} & \multicolumn{2}{c|}{\textbf{Pose Generation}}& \multicolumn{5}{c}{\textbf{Image Synthesis}} \\
    \hline
     &\textbf{MSE}\ $ (\downarrow)$ & \textbf{Var}\ $ (\uparrow)$ & \textbf{SSIM}\ $  (\uparrow)$ & \textbf{LPIPS}\ $ (\downarrow)$ & \textbf{FID}\ $ (\downarrow)$ & \textbf{Aes}\ $ (\uparrow)$ & \textbf{Aes-Qwen}\ $ (\uparrow)$ \\
    \hline
    
    WGAN-LP + DPTN~\cite{zhang2022exploring} & 286.9 & 172.2 & 0.7112 & 0.1931 & 11.387 & 3.91 & 4.25 \\
    
    UNet-T2H + PIDM~\cite{bhunia2023person} & 278.4 & 233.4 & 0.7312 & 0.1678 & 6.3671 & 3.84 & 4.89 \\

    GUNet + CFLD~\cite{lu2024coarse} & 262.2 & 244.5 & 0.7378 & 0.1519 & 6.8040 & 3.97 & 5.35 \\
    GUNet + PCDMs~\cite{shen2023advancing}  & 262.2 & 244.5 & 0.7444 & 0.1365 & 7.4734 & 4.45 & 5.12 \\

    \hline
    
    \rowcolor{gray!10}
    \textbf{FashionPose (Ours)}  & \textbf{243.8} & \textbf{251.7} & \textbf{0.7603} & \textbf{0.1221} & \textbf{5.6690} & \textbf{4.79} & \textbf{5.98} \\
    \bottomrule
    \end{tabular}%
    }
\vspace{-0.3cm}
\end{table*}

\section{Methods}

\noindent\textbf{Overview.}
We position FashionPose within the task of text-driven human image synthesis with Identity Preservation. Unlike generic text-to-image generation where subjects are synthesized from scratch, practical fashion e-commerce operates under a constrained paradigm: the garment and model identity are strictly given (provided via a source image $I_s$), while the objective is to vary the visual presentation, specifically pose and environmental context, through natural language. To address this, FashionPose is designed as a unified framework that reconciles abstract linguistic descriptions with specific visual identities. Our goal is to leverage natural language as the primary interface to jointly control the geometric posture and environmental illumination of a synthesized subject, while rigorously preserving the identity and garment details from a reference source.
As illustrated in Figure~\ref{fig:model_architecture}, our framework operates via hierarchical semantic grounding, where textual descriptions are first projected into a geometric latent space to establish spatial anchors. These anchors subsequently govern the identity-consistent synthesis and act as the basis for joint photometric modulation.
First, we bridge the cross-modal gap by translating the input text directly into an explicit geometric skeleton with joint visibility. Unlike template-based methods, we employ a semantic-driven pose generator that is driven by a bidirectional contrastive alignment mechanism. This design ensures that the predicted body structure is not only physically plausible but also semantically aligned with the specific action nuances described in the language.

\subsection{Bidirectional Contrastive Alignment}
\label{ssec:t2p}

To bridge the gap between abstract linguistic concepts and explicit body geometry, we propose a semantic-driven generation framework. As shown in Figure~\ref{fig:model_architecture}, this stage maps a caption $C$ to a target pose $P_t$ with visibility $\mathbf v$. The core innovation lies in our bidirectional contrastive alignment, which enforces a shared semantic manifold between text and pose representations, ensuring the generated skeletons faithfully reflect action-specific nuances.

The framework consists of two lightweight modules. First, the text conditioning adapter (TCA) projects frozen CLIP token features into a pose-aligned latent space, aggregating them into a global context vector $\mathbf z$. Second, the visibility-aware pose regressor (VPR) decodes $\mathbf z$ into 2D joint coordinates $P=\{\mathbf P_i\}$ and visibility scores $\mathbf v=\{v_i\}$. This decoupled design allows the model to hallucinate plausible structures even for occluded joints.

Standard regression often leads to "mean pose" artifacts. To solve this, we explicitly align the semantic space of the input text with the geometric space of the output pose. We project the text representation and the generated pose into a normalized embedding space via heads $\phi_{\text{text}}$ and $\phi_{\text{pose}}$.
For a batch of $B$ pairs, we maximize the similarity $s_{pq} = \langle \phi_{\text{text}}(C_p), \phi_{\text{pose}}(P_q) \rangle / \tau$ between matched pairs while suppressing mismatched ones in both directions. The alignment objective is formulated as:
\begin{equation}
L_{\mathrm{con}}=
-\frac1{2B}\sum_{p=1}^{B}
\left[\log\frac{e^{s_{pp}}}{\sum_{q} e^{s_{pq}}}
+\log\frac{e^{s_{pp}}}{\sum_{q} e^{s_{qp}}}\right].
\end{equation}
This mechanism serves as a semantic anchor, ensuring that fine-grained descriptions (e.g., "crossed legs" vs. "standing") map to precise geometric configurations.

We complement semantic alignment with explicit geometric supervision. First, we supervise the joint coordinates via MSE for visible parts ($L_{\mathrm{coord}}$) while regularizing invisible ones ($L_{\mathrm{inv}}$) to prevent drifting:
\begin{equation}
L_{\mathrm{coord}} = \frac{\sum_{i} m_i \|\hat{\mathbf P}_i - \mathbf P_i\|^2}{\sum_{i} m_i+\epsilon}, \quad
L_{\mathrm{inv}} = \frac{\sum_{i} (1-v_i)\|\hat{\mathbf P}_i\|^2}{\sum_{i} (1-v_i)+\epsilon}.
\end{equation}
Second, the visibility state is learned via a standard binary cross-entropy (BCE) loss:
\begin{equation}
L_{\mathrm{vis}} = -\frac{1}{|\mathcal J|} \sum_{i}\left[v_i\log\hat v_i+(1-v_i)\log(1-\hat v_i)\right].
\end{equation}
Finally, the total objective is 
\begin{equation}
    L_{\text{T2P}} = L_{\mathrm{coord}} + L_{\mathrm{vis}} + \lambda_{\text{inv}}L_{\mathrm{inv}} + \lambda_{\text{con}}L_{\mathrm{con}}.
\end{equation}

\subsection{Identity-Anchored Synthesis}
\label{ssec:p2i}

Given the target skeleton $P_t$ derived from textual semantics, this module aims to synthesize a high-fidelity image $I_p$ that strictly aligns with $P_t$ while rigorously preserving the identity and garment details from the source image $I_s$. 

Following the framework, we use frozen encoders to extract conditions from the two input branches (source image $I_s$ \& mask $M_s$) and (source pose $P_s$ \& target pose $P_t$), and fuse them by element-wise addition:
\begin{equation}
\mathbf F
=
\mathrm{Enc}_{\text{img}}(I_s,M_s)
+
\mathrm{Enc}_{\text{pose}}(P_s,P_t).
\end{equation}
To complement the spatial layout with robust identity cues, we parallelly extract token-level appearance embeddings $\mathbf A = \mathrm{DINO}(I_s)$ via a frozen DINO encoder. These embeddings serve as semantic anchors within the diffusion backbone. Consequently, the denoiser synthesizes the target image $I_p$ by predicting the noise residual $\hat\epsilon = \epsilon_{\theta_p}(x_t, t; \mathbf F, \mathbf A)$, conditioned jointly on the fused spatial map and the appearance guidance. We optimize the standard noise-prediction loss with an additional mask-based anchoring term on identity-critical regions:
\begin{equation}
L_{\text{P2I}}
=
\mathbb E\big[\|\epsilon-\hat\epsilon\|_2^2\big]
+
\lambda_{\text{id}}\big\|M_s\odot(\hat I_p-I_s)\big\|_1,
\end{equation}
where $\hat I_p$ denotes the predicted clean image and $\lambda_{\text{id}}\ge 0$ controls the anchoring strength.

\subsection{Prompt-Conditioned Relighting}
\label{ssec:relight}

While the previous module ensures structural fidelity, it operates under neutral or studio-like illumination. To bridge the gap between the synthesized subject and the complex environmental context described in the text, we introduce a prompt-conditioned relighting module. As illustrated in Figure~\ref{fig:model_architecture}, this module aims to harmonize the shading and shadow effects of the pose-aligned image $I_p$ with the photometric semantics of the prompt $T$, all while rigorously preserving the identity and garment details established previously.

To ensure that the relighting process modifies only the illumination without altering the structure or texture, we utilize a strong content prior. We extract a high-level content representation $\mathbf z_p$ from the pose image $I_p$ using an image encoder:
\begin{equation}
\mathbf z_p=\mathrm{Enc}_{\text{img}}(I_p).
\end{equation}
This representation serves as a structural anchor. During the diffusion process, rather than starting from pure noise, we concatenate this content latent $\mathbf z_p$ with the noisy latent $\boldsymbol{\eta}_t$ at each timestep $t$. This input formulation $\mathbf x_t = \mathrm{concat}(\mathbf z_p, \boldsymbol{\eta}_t)$ forces the denoiser to strictly respect the spatial layout and identity of the input subject.

The desired lighting condition is injected via the textual prompt $T$. We encode $T$ using a frozen CLIP text encoder to guide the denoiser $\epsilon_{\theta_r}$. The model is trained to predict the added noise under this dual conditioning (structural content + photometric prompt) via the objective:
\begin{equation}
L_{\text{relight}}
=\mathbb E_{t,\boldsymbol{\epsilon}}
\Big[
\big|
\boldsymbol{\epsilon}-\epsilon_{\theta_r}\big(\mathbf x_t,\ t;\ \mathrm{Enc}^{\mathrm{CLIP}}_{\text{text}}(T)\big)
\big|_2^2
\Big].
\end{equation}

\section{Experiments And Analysis}

\subsection{Implementation Details}

\noindent\textbf{Metrics.}
We employ a comprehensive set of metrics tailored to each evaluation phase.
For the end-to-end framework, we assess pose generation using MSE~\cite{jahne2005digital} for coordinate accuracy and Var for diversity. Image synthesis quality is evaluated via standard SSIM~\cite{wang2004image}, PSNR~\cite{jahne2005digital}, and FID~\cite{heusel2017gans}, complemented by Aes and Aes-Qwen. Aes is a widely adopted embedding-based scorer developed by LAION~\cite{schuhmann2022laion}. Aes-Qwen is an aesthetic quality score ranging from 0 to 10 rated by Qwen-VL~\cite{bai2023qwenvlversatilevisionlanguagemodel}.
For the relighting experiments, we report CVS~\cite{radford2021learning} for semantic alignment, FIS~\cite{deng2019arcface} for identity preservation, and LS~\cite{gower1975generalized} to evaluate lighting consistency.
Finally, for ablation studies focusing on geometric precision, we utilize MPJPE~\cite{ionescu2013human3} and PCK~\cite{yang2012articulated} for localization error, alongside mAP~\cite{everingham2010pascal} for visibility prediction. To complement these objective assessments, we conduct a user study employing R2G (real-to-generated) and G2R (generated-to-real) metrics to evaluate perceptual fidelity and realism. Additionally, subjective aesthetic quality is directly rated by human participants, providing a subjective counterpart to the automated Aes and Aes-Qwen metrics.



\noindent\textbf{Datasets.} To evaluate FashionPose, we adopt a two-tiered strategy that first validates end-to-end capabilities, followed by component-wise benchmarking. We begin by evaluating unified text-to-pose generation performance on our proposed PoseCap dataset, which contains over 40k caption-keypoint pairs. Subsequently, to verify individual module robustness, we evaluate pose-guided synthesis on the standard DeepFashion~\cite{liu2016deepfashion} and Market-1501~\cite{Zheng_2015_ICCV}.

\begin{figure}[t]
\centering
\includegraphics[width=1\linewidth]{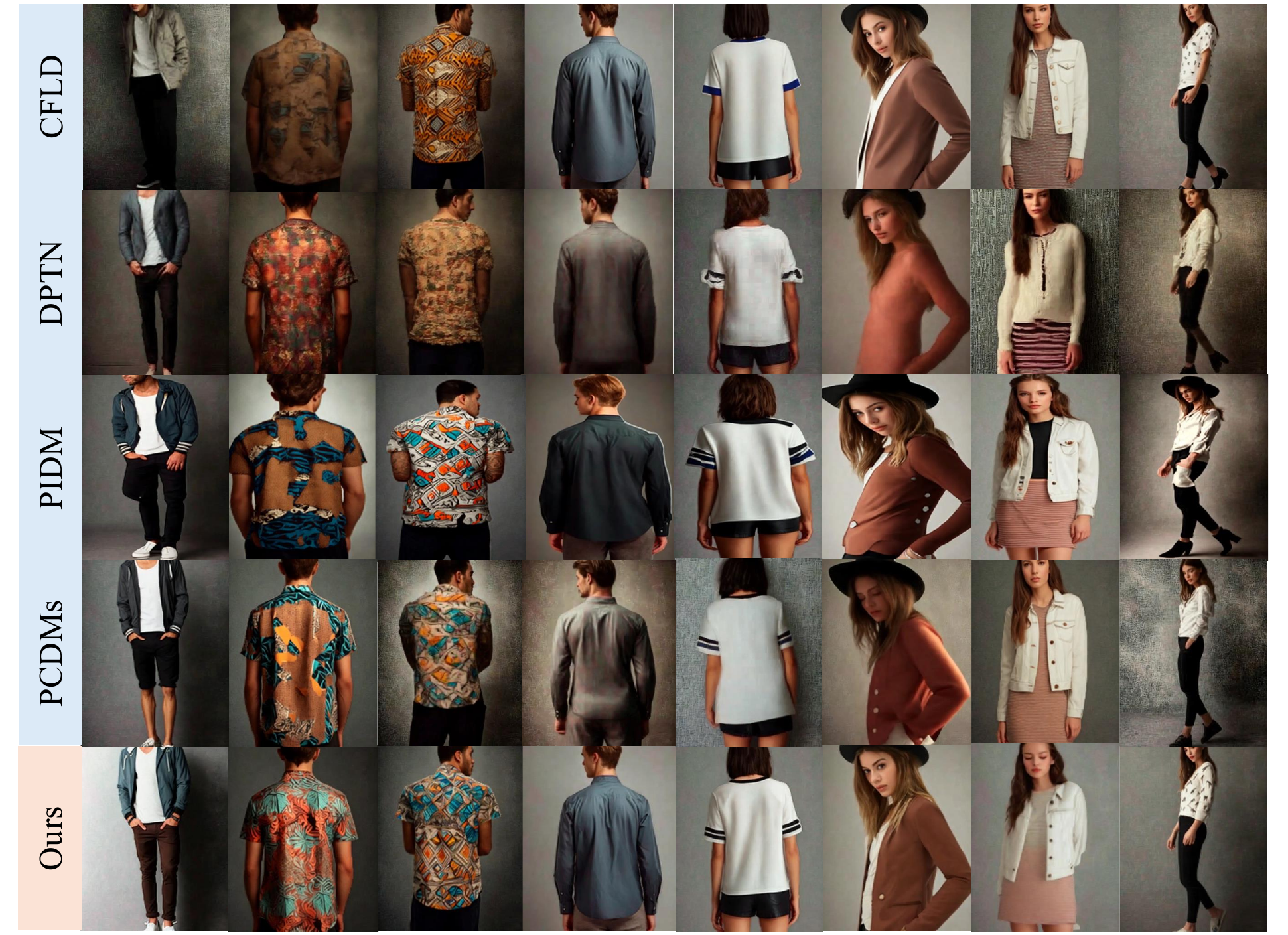}
\caption{\textbf{Qualitative comparison on the DeepFashion test set.} Compared with SOTA baselines (top line), our method (bottom line) synthesizes significantly sharper garment textures and maintains structural coherence even under large pose deformations, effectively reconstructing occluded regions.}
\label{fig:compare}
\vspace{-0.3cm}
\end{figure}

\noindent\textbf{Hyperparameters.} We keep the CLIP text encoder frozen and use a Transformer with hidden dimension 640, 4 layers, 4 attention heads, and dropout $p=0.05$. We adopt the best-performing objective-related settings from the sweeps, with $\lambda_{\mathrm{inv}}=0.25$ and $\lambda_{\mathrm{con}}=0.05$. For the contrastive alignment, we set the temperature $\tau=0.07$ and use a batch size of $B=64$. Optimization is performed with AdamW at a learning rate of 1e-4. For pose-conditioned synthesis, we employ Stable Diffusion v1.5 as the backbone operating at $512 \times 512$ resolution. We train the model using the standard noise-prediction objective augmented with an identity-anchoring term ($\lambda_{\text{id}}$) with a weight of 0.05.

\subsection{Compare With SOTA Methods}
\noindent\textbf{Quantitative Results.}
To demonstrate the effectiveness of our end-to-end framework and its superiority over existing multi-stage solutions, including strong baselines enhanced with modern relighting tools like IC-Light~\cite{zhang2025scaling}, we conducted a rigorous quantitative evaluation on the PoseCap dataset. We constructed composite baselines using state-of-the-art methods for each stage (e.g., GUNet~\cite{liang2024gunetgraphconvolutionalnetwork} for pose generation and PCDMs~\cite{shen2023advancing}/PIDM~\cite{bhunia2023person} for image synthesis) and evaluated performance across structural accuracy (MSE, Var) and perceptual realism (FID, Aes-S). The results indicate that our method achieves State-of-the-Art performance across all metrics, effectively bridging the gap between structural precision and textural realism. Specifically, in the pose generation stage, our model significantly reduces keypoint error (MSE of 243.8 vs. GUNet's 262.2) while maintaining superior diversity (Var of 251.7). In the image synthesis stage, leveraging precise pose guidance, our full framework achieves a remarkable FID of 5.6690, representing a 11.0\% improvement over the strongest baseline combination (UNet-T2H~\cite{liang2024gunetgraphconvolutionalnetwork} + PIDM), alongside the highest SSIM of 0.7603. Notably, on the aesthetic score (Aes-S), our method achieves 5.98, outperforming even the baselines equipped with IC-Light for relighting (which peaked at 5.35). In summary, these quantitative results strongly validate that our proposed architecture not only accurately translates text into plausible human structures but also directly generates high-fidelity, aesthetically pleasing fashion images without relying on external post-processing tools.

\noindent\textbf{Qualitative Results.} 
To qualitatively assess visual fidelity, we compare our method against leading pose-guided synthesis baselines, as shown in Figure~\ref{fig:compare}. The visual comparisons demonstrate that our approach generates the most photorealistic results with superior detail preservation. Specifically, in Column 1, our method maintains strict garment consistency with sharp boundaries even under significant pose transformations. This superiority is further evident in Columns 2–4, where we faithfully reconstruct complex textures and high-frequency patterns; in contrast, competitors like DPTN tend to blur these intricate details, while CFLD often degrades material authenticity. Moreover, Columns 5 and 6 highlight our precision in synthesizing fine-grained attributes, such as cuffs and buttons, which appear distinct and realistic compared to the compromised detail quality observed in the baselines. Consequently, our method achieves a superior balance between structural alignment and textural fidelity, avoiding the artifacts common in existing approaches.

\begin{figure}[t]
\centering
\includegraphics[width=1\linewidth]{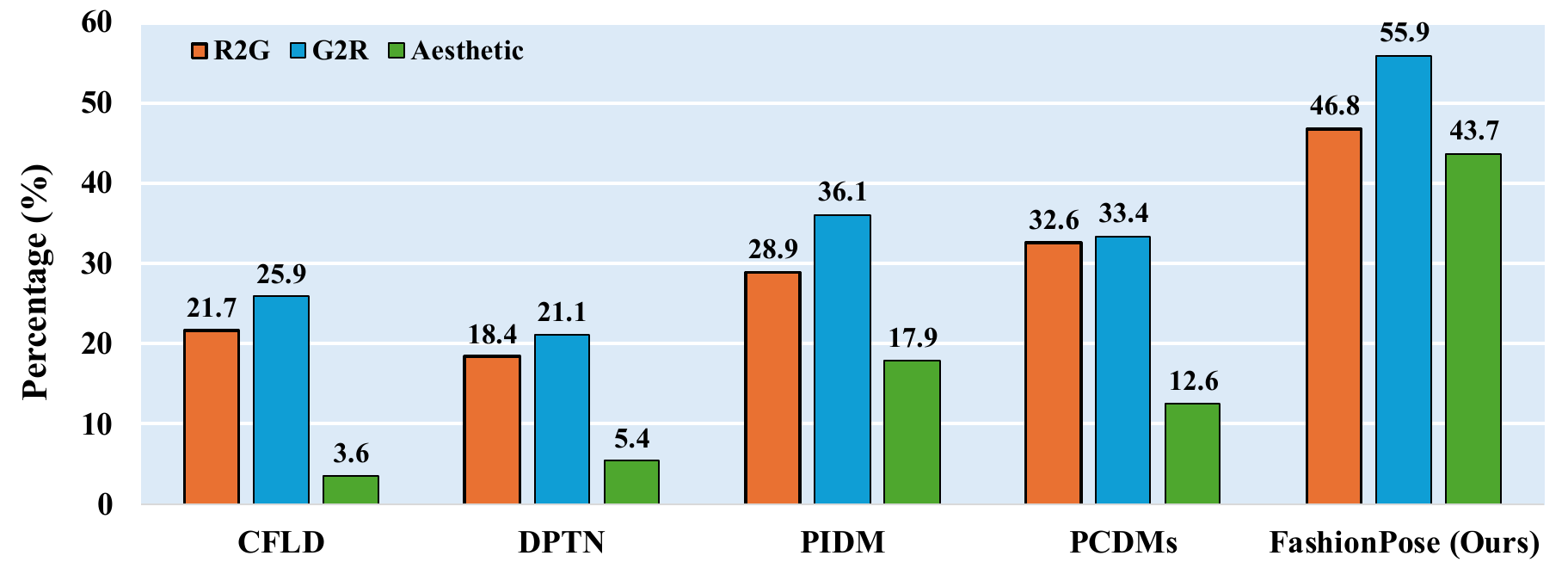}
\caption{\textbf{User study results on PoseCap in terms of
R2G, G2R, and Aesthetic metric.} Higher values in these three
metrics indicate better performance.}
\label{fig:user}
\vspace{-0.3cm}
\end{figure}

\begin{table}[t]
\centering
\caption{\textbf{Quantitative comparison of text-guided 2D pose generation with $K{=}10$ samples per text.} MSE measures keypoint accuracy and Var measures sampling diversity.}
\label{tab:gunet_quant1_with_ours}
\begin{tabular}{lcc}
\hline
\rowcolor{gray!20}
Methods & MSE $(\downarrow)$ & Var $(\uparrow)$ \\
\hline
GUNet~\cite{liang2024gunetgraphconvolutionalnetwork} & 262.2 & 244.5 \\
UNet-T2H~\cite{liang2024gunetgraphconvolutionalnetwork} & 278.4 & 233.4 \\
WGAN-LP~\cite{zhang2020adversarialsynthesishumanpose} & 286.9 & 172.2 \\
WGAN-LP R~\cite{zhang2020adversarialsynthesishumanpose} & 290.8 & 131.2 \\
\hline
\rowcolor{gray!10}\textbf{FashionPose (Ours)} & \textbf{251.3} & \textbf{249.2} \\
\hline
\end{tabular}
\vspace{-0.3cm}
\end{table}

\noindent\textbf{User Study.} To evaluate the perceptual differences in pose consistency and overall visual quality among different methods under language-driven conditions, we conduct a user study on the PoseCap dataset. We recruit 30 participants, who are asked to compare results generated by different methods given the same textual prompt and reference image. Each comparison is assessed along three dimensions: R2G consistency, G2R recoverability, and overall aesthetic quality. The results show that FashionPose achieves the highest preference ratios across all three criteria. Specifically, FashionPose attains preference ratios of 46.8\%, 55.9\%, and 43.7\% on R2G, G2R, and aesthetic quality, respectively, demonstrating consistent and significant improvements over the strongest baseline. These findings further validate the importance of adopting language as a unified interface and jointly modeling geometric and photometric factors to enhance perceptual quality in practical synthesis scenarios.

\subsection{Ablation Studies}

\noindent\textbf{Text to Pose Generation.} To quantitatively assess the fidelity and generative diversity of our text-guided 2D pose generation, we conducted a controlled study generating $K{=}10$ samples per prompt, evaluating Keypoint Accuracy (MSE, lower is better) and Sampling Diversity (Var, higher is better). As shown in Table~\ref{tab:gunet_quant1_with_ours}, the results demonstrate that FashionPose achieves the best overall performance, effectively balancing precision with variation. Specifically, in terms of structural accuracy, we attained the lowest MSE of 251.3, significantly outperforming the runner-up GUNet (262.2) and establishing a substantial margin against UNet-T2H (278.4). Crucially, this precision does not come at the cost of diversity; we simultaneously achieved the highest Variance of 249.2, surpassing GUNet (244.5) and far exceeding the WGAN-LP baselines (172.2 and 131.2). 
Ablation results confirm the indispensability of our decoupled design; while the base synthesis model ensures realism, only the integration of our relighting and alignment modules achieves true semantic-photometric harmony.

\begin{table}[t]
\centering
\caption{\textbf{Quantitative comparison with several pose image generation SOTA models.} Comparison using SSIM, LPIPS, and FID.}
\renewcommand{\arraystretch}{1}
\resizebox{\linewidth}{!}{%
    \begin{tabular}{c|l|c|c|c}
    \hline
    \rowcolor{gray!20}
    \textbf{Dataset} & \textbf{Methods} & \textbf{SSIM}$(\uparrow)$ & \textbf{LPIPS}$(\downarrow)$ & \textbf{FID}$(\downarrow)$ \\
    \hline

    \multirow{7}{*}{\begin{tabular}{c}
    DeepFashion\\
    (256$\times$176)
    \end{tabular}}
    & CASD            & 0.7248 & 0.1936 & 11.373 \\
    & PoCoLD          & 0.7310 & 0.1642 & 8.0667 \\
    & PIDM            & 0.7312 & 0.1678 & 6.3671 \\
    & CFLD            & 0.7378 & 0.1519 & 6.8040 \\
    & PCDMs           & 0.7444 & 0.1365 & 7.4734 \\
    & IMAGPose        & 0.7561 & 0.1284 & 5.8738 \\
    & \textbf{Ours} & \textbf{0.7603} & \textbf{0.1217} & \textbf{4.6034} \\
    \hline

    \multirow{6}{*}{\begin{tabular}{c}
    DeepFashion\\
    (512$\times$352)
    \end{tabular}}
    & PIDM            & 0.7419 & 0.1768 & 5.8365 \\
    & PoCoLD          & 0.7430 & 0.1920 & 8.4163 \\
    & CFLD            & 0.7478 & 0.1819 & 7.1490 \\
    & PCDMs           & 0.7601 & 0.1475 & 7.5519 \\
    & IMAGPose        & 0.7718 & 0.1396 & 5.6298 \\
    & \textbf{Ours} & \textbf{0.7809} & \textbf{0.1304} & \textbf{4.8713} \\
    \hline

    \multirow{5}{*}{\begin{tabular}{c}
    Market-1501\\
    (128$\times$64)
    \end{tabular}}
    & DPTN            & 0.2854 & 0.2711 & 18.995 \\
    & PIDM            & 0.3054 & 0.2415 & 14.451 \\
    & PCDMs           & 0.3169 & 0.2238 & 13.897 \\
    & IMAGPose        & 0.3282 & 0.2104 & 12.659 \\
    & \textbf{Ours} & \textbf{0.3348} & \textbf{0.2079} & \textbf{11.633} \\
    \hline

    \end{tabular}%
}
\label{quantitative_comparison}
\vspace{-0.3cm}
\end{table}

\noindent\textbf{Pose Image Generation.} To comprehensively evaluate the generation quality and geometric consistency of our framework, we conducted comparative experiments against state-of-the-art models on the DeepFashion (at resolutions $256\times176$ and $512\times352$) and Market-1501 datasets. As shown in Table~\ref{quantitative_comparison}, the results demonstrate that our method consistently achieves the best performance across SSIM, LPIPS, and FID metrics. Specifically, on the $256\times176$ DeepFashion benchmark, our model outperforms the strong competitor IMAGPose, improving SSIM to 0.7603 and significantly reducing FID from 5.8738 to 4.6034. This advantage is even more pronounced against PCDMs, which lags behind with an FID of 7.4734. Similarly, in the high-resolution setting ($512\times352$), we maintain a clear lead, lowering LPIPS to 0.1304 compared to IMAGPose's 0.1396. On the challenging Market-1501 dataset, we further validate our robustness by surpassing IMAGPose, achieving a lower FID of 11.633 versus 12.659. Consequently, the consistent reduction in perceptual error alongside higher structural similarity confirms that our explicit skeleton regression strategy effectively enhances geometric alignment without compromising visual realism.

\begin{table}[t]
  \centering
  \caption{\textbf{Quantitative comparison for text-guided portrait relighting against representative instruction-based baselines.} CVS measures CLIP-vision similarity, FIS measures ArcFace-based identity similarity, and LS is a LLaVA-based lighting adjustment score.}
  \vspace{-0.2cm}
  \label{tab:relight_sota}
  \resizebox{\linewidth}{!}{
  \begin{tabular}{l|ccccc}
    \hline
    \rowcolor{gray!20}
    Method & SSIM $(\uparrow)$ & LPIPS $(\downarrow)$ & CVS $(\uparrow)$ & FIS $(\uparrow)$ & LS $(\uparrow)$ \\
    \hline
    IP2P        & 0.408 & 0.531 & 0.646 & 0.008  & 3.6 \\
    GLIDE       & 0.464 & 0.525 & 0.643 & -0.969 & 3.6 \\
    MGIE        & 0.415 & 0.464 & 0.802 & 0.465  & 3.3 \\
    Text2Relight& 0.546 & 0.401 & 0.886 & 0.584  & 3.8 \\
    IC-Light    & 0.562 & 0.393 & 0.895 & 0.591  & 4.0 \\
    \hline
    \textbf{Ours}        & \textbf{0.572} & \textbf{0.388} & \textbf{0.911} & \textbf{0.602} & \textbf{4.2} \\
    \hline
  \end{tabular}}
\end{table}

\noindent\textbf{Prompt-Conditioned Relighting.} To rigorously evaluate the effectiveness of our prompt-conditioned relighting module, we conducted a unified quantitative assessment against existing state-of-the-art methods. The results indicate that our approach ranks first across all evaluated metrics (Table~\ref{tab:relight_sota}). Specifically, compared with the strongest baseline, Text2Relight, our model significantly improves structural fidelity, boosting SSIM to 0.572 (up from 0.546) while reducing perceptual error (LPIPS) to 0.388. Beyond image quality, we also achieve superior semantic control, increasing the CVS to 0.911 and LS to 4.0, surpassing the baseline's 0.886 and 3.8, respectively. Crucially, this lighting manipulation does not compromise identity; we raise the FIS from 0.584 to 0.602. These improvements demonstrate our method's ability to synthesize complex lighting effects that are both semantically accurate and faithful to the subject's identity.

\begin{table}[t]
    \centering
    \caption{\textbf{Robustness to Pose Perturbations.} Quantitative evaluation of SSIM under varying levels of Gaussian noise ($\sigma$) injected into stage 1.}
    \label{tab:noise_robustness}
    \begin{tabular}{c|c}
        \hline
        \rowcolor{gray!20}
        \textbf{Noise Level ($\sigma$)}  & \textbf{SSIM} ($\uparrow$)  \\
        \hline
        0 &  \textbf{0.7603}  \\
        2 &  0.7595  \\
        4 &  0.7571 \\
        6 &  0.7568  \\
        \hline
    \end{tabular}
    \vspace{-0.3cm}
\end{table}

\begin{figure}[t]
    \centering
    \includegraphics[width=1\linewidth]{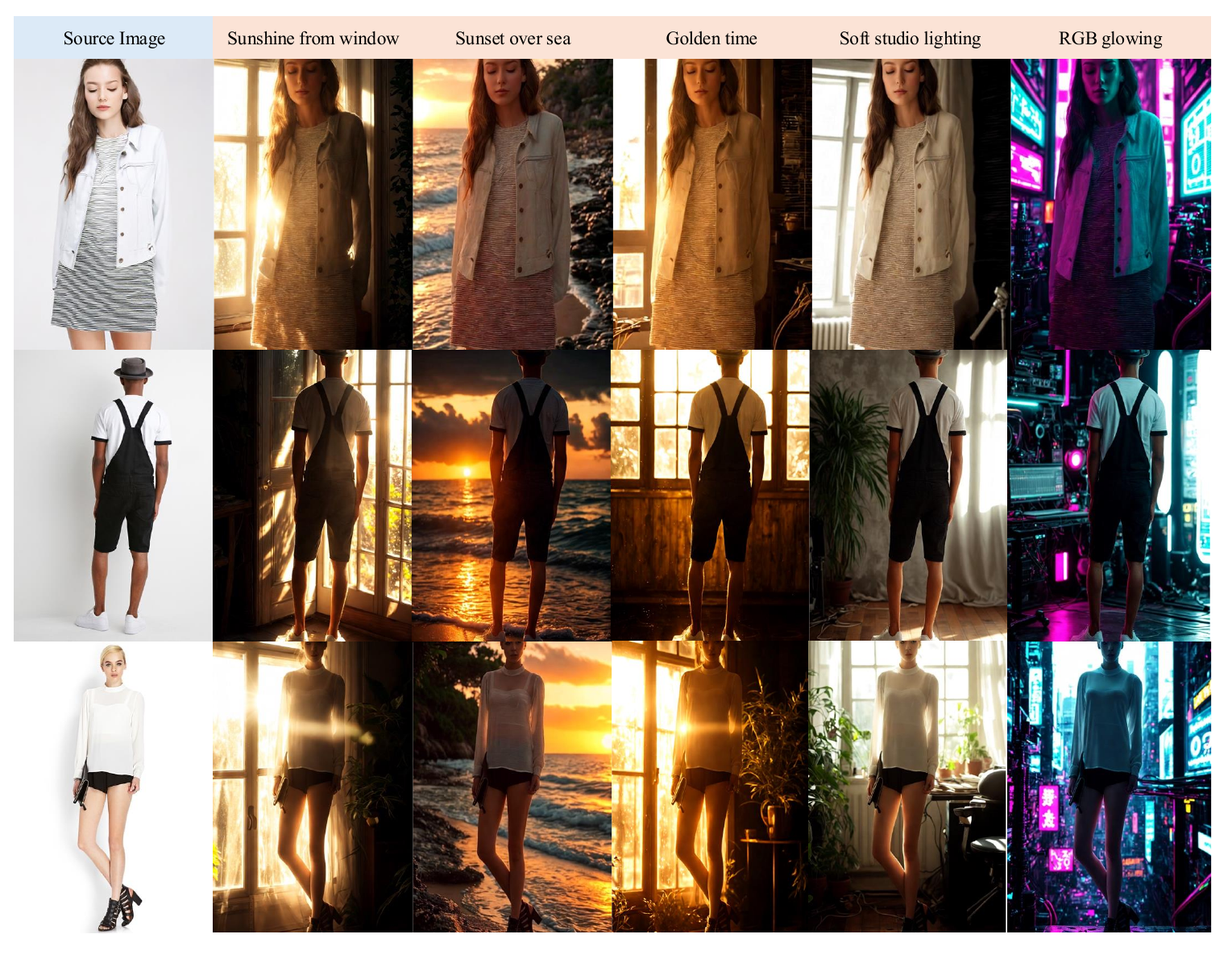}
    \caption{\textbf{Text-conditioned relighting.} Unlike methods limited to calibrated setups, FashionPose enables photometric control in diverse non-studio conditions. }
    \label{fig:light}
    \vspace{-0.3cm}
\end{figure}

\begin{figure*}[t]
  \centering
  \includegraphics[width=1\linewidth]{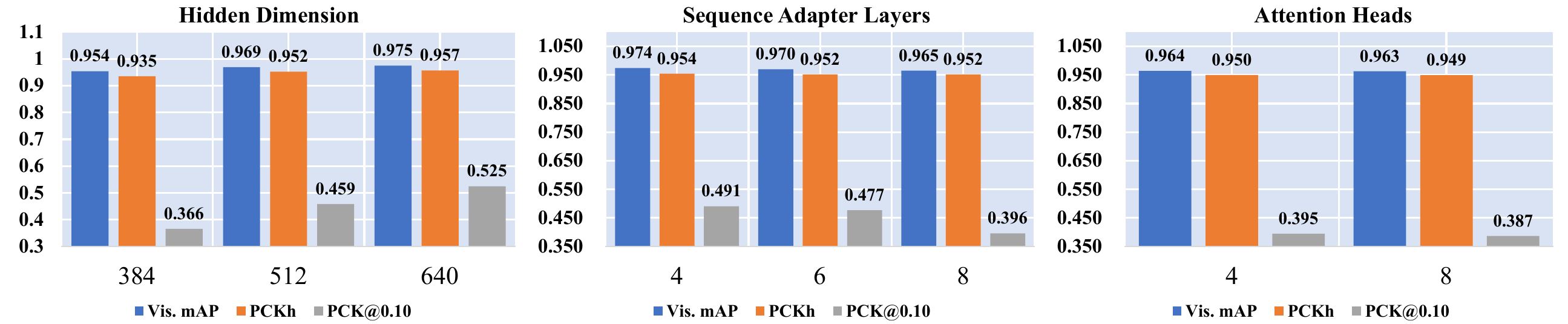}
  \caption{\textbf{One-factor-at-a-time sensitivity of architectural on the PoseCap.} Results comparing different parameters for hidden dimensions, sequence adapter layers, and attention heads.}
  \label{fig:hyper1}
\end{figure*}

\noindent\textbf{Impact of Geometric Noise on Generation Fidelity.}
To evaluate the structural robustness of our framework against geometric imperfections, we designed a sensitivity experiment by injecting five levels of Gaussian noise ($\sigma \in \{0, 2, 4, 6\}$) into the generated keypoints, concluding that FashionPose exhibits significant resilience to pose jitter. Specifically, even under a noise intensity of $\sigma=6$, the SSIM score demonstrated only a negligible degradation, decreasing smoothly from the baseline of 0.7603 to 0.7568. Thus, this proves that our framework effectively mitigates upstream geometric errors, leveraging strong visual priors to maintain structural fidelity despite input perturbations.

\subsection{Application} 
While conventional relighting paradigms are largely confined to controlled studio setups with simplified backdrops, our framework transcends these boundaries, enabling high-fidelity illumination synthesis in unconstrained, complex real-world environments. As visualized in Figure~\ref{fig:light}, we demonstrate this versatility across a diverse spectrum of unseen prompts, ranging from atmospheric natural lighting (e.g., \textit{sunshine from window}, \textit{golden hour}) to stylized artistic effects (e.g., \textit{RGB glow}). In each scenario, our module exhibits precise control over photometric attributes modulating directionality, color temperature, and intensity while rigorously preserving the subject's identity and intricate garment details. These results validate a paradigm shift: a single natural-language interface now suffices to orchestrate sophisticated photometric control, bridging the gap between professional studio lighting and dynamic environmental rendering. Moreover, our method demonstrates strong generalization on unconstrained \textbf{in-the-wild} images; please refer to the Appendix for additional visualizations.

\subsection{Hyperparameter Sensitive Analysis}

\noindent\textbf{Variant Structures.} We analyze the sensitivity of key structural hyperparameters by varying one factor at a time while keeping all other training and inference settings fixed. Increasing the hidden dimension from 384 to 640 yields consistent improvements, with Vis.\ mAP rising from 0.954 to 0.975, PCKh from 0.935 to 0.957, and PCK@0.10 from 0.366 to 0.525, indicating that larger representations particularly benefit fine-grained localization. In contrast, the number of sequence adapter layers exhibits a clear optimum: 4 layers achieves the best overall results with Vis.\ mAP 0.974, PCKh 0.954, and PCK@0.10 0.491, while deeper adapters degrade performance, most notably on PCK@0.10, which drops to 0.396 at 8 layers. Finally, the attention head count has limited impact in the tested range, with 4 heads slightly outperforming 8 heads across all metrics, suggesting robustness to this choice. Based on these observations, we use dimension 640, 4 adapter layers, and 4 heads as the default configuration.

\begin{table}[t]
\centering
\caption{\textbf{One-factor-at-a-time sensitivity of $\lambda$ on the PoseCap.} }
\label{tab:hyper_ablation}
    \begin{tabular}{cc|ccc} 
    \hline
    \textbf{Param.} & \textbf{Value} & \textbf{Vis. mAP} & \textbf{PCK@0.1} & \textbf{PCKh} \\
    \hline
    \multirow{3}{*}{$\lambda_{\text{inv}}$} 
      & 0.25 & \textbf{0.968} & \textbf{0.486} & \textbf{0.954} \\
      & 0.50 & 0.967 & 0.441 & 0.950 \\
      & 1.00 & 0.930 & 0.370 & 0.949 \\
    \hline
    \multirow{3}{*}{$\lambda_{\text{con}}$} 
      & 0.05 & \textbf{0.970} & \textbf{0.457} & \textbf{0.952} \\
      & 0.10 & 0.966 & 0.392 & 0.952 \\
      & 0.20 & 0.965 & 0.410 & 0.951 \\
    \hline
    \end{tabular}
    \vspace{-0.5cm}
\end{table}

\noindent\textbf{Hyperparameter Analysis.} We vary one coefficient at a time on the validation set while keeping all other settings fixed (Table~\ref{tab:hyper_ablation}). For the invisible-joint regularizer, $\lambda_{\text{inv}}=0.25$ achieves the best overall performance (PCKh 0.954, PCK@0.10 0.486, Vis.\ mAP 0.968). Increasing $\lambda_{\text{inv}}$ to 0.50 slightly decreases PCKh (0.950) and notably drops the strict metric PCK@0.10 (0.441) while Vis.\ mAP stays similar (0.967). Further increasing it to 1.00 results in significant degradation in Vis.\ mAP (0.930) and PCK@0.10 (0.370), suggesting overly strong regularization harms fine-grained localization and overall accuracy. For structure--semantic coupling, $\lambda_{\text{con}}=0.05$ yields the best Vis.\ mAP (0.970) and PCK@0.10 (0.457) with PCKh 0.952. Larger weights reduce both Vis.\ mAP (0.966/0.965) and PCK@0.10 (0.392/0.410), while PCKh remains nearly unchanged (0.952 to 0.951), indicating that excessive coupling may weaken coordinate-level supervision. We thus set $\lambda_{\text{inv}}=0.25$ and $\lambda_{\text{con}}=0.05$.

\begin{figure}[t]
    \centering
    \includegraphics[width=0.9\linewidth]{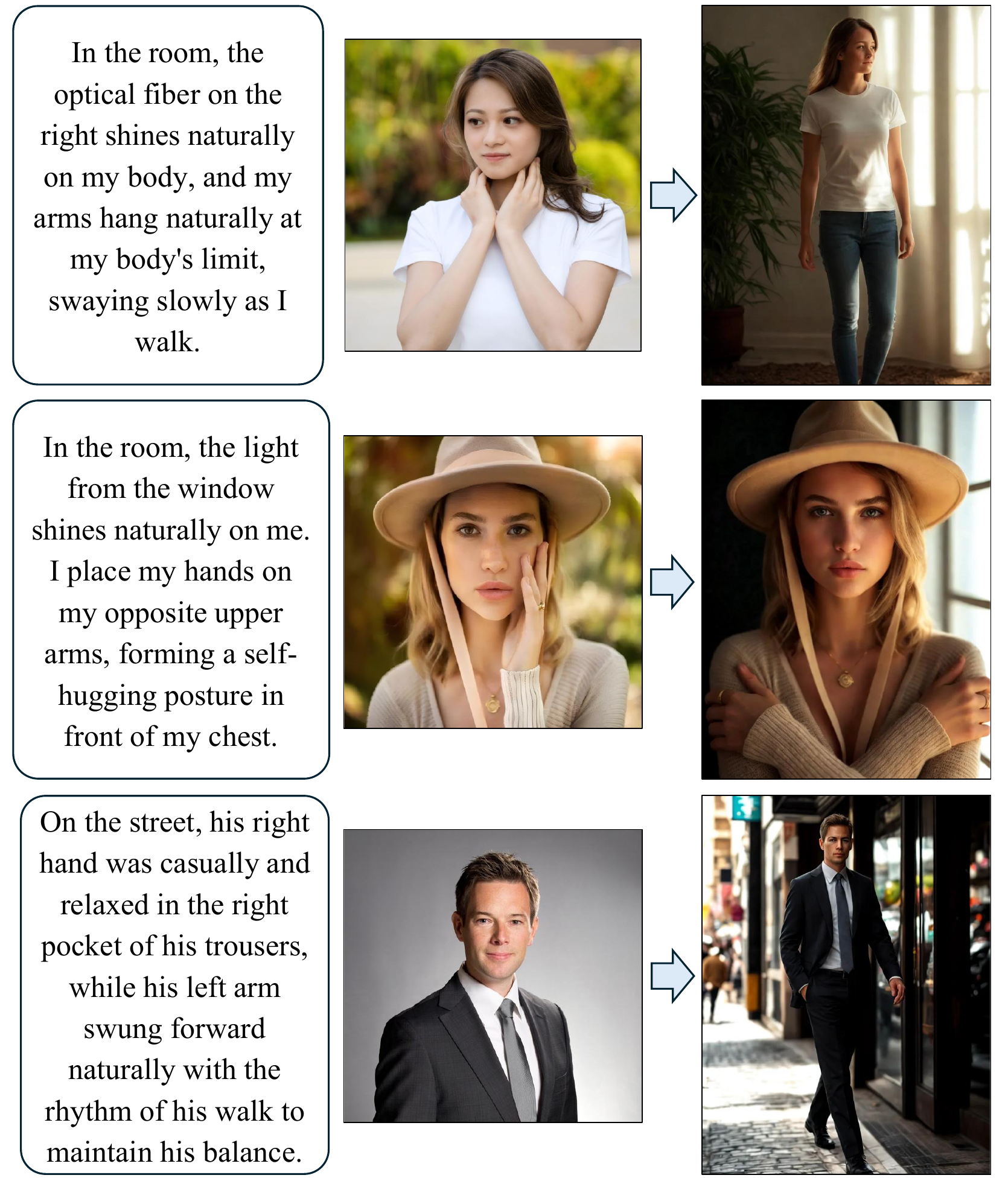}
    \caption{\textbf{Illustration of the PoseCap dataset structure.} Each data sample consists of a high-quality image-pose-text triplet. }
    \label{fig:inthewild}
    \vspace{-0.3cm}
\end{figure}

\noindent\textbf{More Result In-The-Wild.} \label{more}
To assess the generalization capability of FashionPose beyond the semi-structured PoseCap domain, we conducted extensive inference on "in-the-wild" images subject to diverse environmental constraints. As visually corroborated in Figure~\ref{fig:inthewild}, beyond exhibiting geometric stability, our model demonstrates precise text-driven photometric control; regardless of whether the prompts specify natural outdoor illumination or complex indoor lighting, the model successfully harmonizes the foreground subject with the target ambient atmosphere. This capability validates that our framework has not merely memorized the training distribution but has acquired a generalized understanding of the 3D interaction among light, surface geometry, and semantic context.

\section{Conclusion} 
In this work, we present FashionPose, a unified framework that makes natural language the primary interface for joint geometric and photometric control in fashion synthesis. We achieved this through a three-stage framework: first, regressing explicit 2D skeletons via bidirectional contrastive alignment to eliminate the need for external templates; second, synthesizing high-fidelity images anchored by identity tokens; and third, applying a prompt-conditioned relighting module to harmonize environmental shading without compromising garment details. To support this paradigm, we introduced PoseCap, a large-scale dataset of over 40k caption–keypoint pairs. Extensive experiments demonstrate that FashionPose delivers superior pose accuracy and physical realism, making it a robust solution for next-generation personalized fashion e-commerce displays.

\bibliographystyle{plain}
\bibliography{ref}

\clearpage
\newpage

\begin{appendices}

\section*{Appendix}
The appendices provide additional details and supplementary materials to support and extend the main paper. Specifically, Appendix~\ref{theo} presents detailed mathematical derivations and theoretical justifications for our core loss functions and conditioning mechanisms, including the bidirectional contrastive alignment and the identity-anchored diffusion objective. Appendix~\ref{dis} offers further discussion on our framework, addressing the rationale behind the cascaded architecture, the model's in-the-wild generalization capabilities, and the training strategy for the prompt-conditioned relighting module. Appendix~\ref{lim} discusses the current limitations of our approach, such as inference latency, and outlines potential avenues for future research. Finally, Appendix~\ref{ethic} concludes with an ethics statement regarding the responsible deployment and potential societal impacts of high-fidelity human image synthesis technologies.

\section{Theoretical Formulations and Optimization Analysis}
\label{theo}
While the main text focuses on the architectural design and empirical validation of FashionPose, this section provides a deeper theoretical analysis of the core loss functions and conditioning mechanisms from the perspectives of information theory, reparameterization, and score-matching.

\subsection{Info-Theoretic Contrastive Alignment}
In the pose generation stage, the bidirectional contrastive alignment loss $L_{con}$ is formulated to map text semantics to physical geometry. Standard regression relies heavily on Mean Squared Error (MSE), which often leads to "mean pose" artifacts. We interpret why $L_{con}$ resolves this from an information-theoretic standpoint.

The bidirectional contrastive loss is inherently a variant of the InfoNCE loss. Let the text embedding space be $X$ and the generated pose embedding space be $Y$. Minimizing $L_{con}$ is mathematically equivalent to maximizing a lower bound on the Mutual Information (MI), $I(X;Y)$, between the text and the pose representations:
\begin{equation}
    I(X;Y) \ge \log(B) - \mathbb{E}_{(x,y)\sim p(x,y)} \left[ -\log \frac{e^{s(x,y)/\tau}}{\frac{1}{B}\sum_{i=1}^B e^{s(x,y_i)/\tau}} \right],
\end{equation}
where $B$ is the batch size and $s(\cdot, \cdot)$ is the cosine similarity. From a gradient perspective, when facing the intrinsic one-to-many ambiguity of text descriptions (e.g., "standing"), an MSE loss pushes the prediction toward the statistical expectation of all possible standing poses, resulting in a blurred or physically implausible average skeleton.

Conversely, because the denominator in Eq. (1) includes the negative samples $\sum_q e^{s_{pq}}$, the gradients of the contrastive loss exert a repulsive force. This pushes the predicted pose away from mismatched structural clusters, forcing the network to converge onto a specific, explicit physical manifold and preserving structural distinctiveness.

\subsection{Reparameterization in Synthesis}
During the identity-anchored synthesis stage, Eq. (6) introduces a mixed diffusion objective $L_{P21}$ that incorporates an $L_1$ penalty in the pixel space. Since the diffusion network $\epsilon_\theta$ operates in the noise space, computing an $L_1$ loss on the image space requires reparameterization through the reverse diffusion process.

At any arbitrary timestep $t$, given the noisy latent $x_t$ and the predicted noise $\hat{\epsilon} = \epsilon_\theta(x_t, t, c)$, the predicted clean image $\hat{I}_P$ (i.e., $\hat{x}_0$) can be derived using the closed-form solution of the forward process:
\begin{equation}
    \hat{I}_P = \frac{x_t - \sqrt{1-\bar{\alpha}_t}\hat{\epsilon}}{\sqrt{\bar{\alpha}_t}},
\end{equation}
where $\bar{\alpha}_t$ is the noise schedule parameter. Substituting this into the pixel-level penalty in Eq. (6) makes the objective fully differentiable with respect to the network parameters $\theta$.

By applying the chain rule, the gradient for the identity-masked region $M_s$ is formulated as:
\begin{equation}
    \nabla_\theta L_{id} = \lambda_{id} M_s \odot \text{sgn}(\hat{I}_P - I_s) \left( -\frac{\sqrt{1-\bar{\alpha}_t}}{\sqrt{\bar{\alpha}_t}} \right) \nabla_\theta \hat{\epsilon}.
\end{equation}
This gradient equation reveals that at low-noise stages ($t \to 0, \bar{\alpha}_t \to 1$), the gradient penalty on the noise predictor is relatively small. However, at high-noise stages ($t \to T, \bar{\alpha}_t \to 0$), the multiplier $-\frac{\sqrt{1-\bar{\alpha}_t}}{\sqrt{\bar{\alpha}_t}}$ is significantly amplified. This dynamic inherently forces the network to strictly align the identity and garment texture with the source image $I_s$ during the early phases of the denoising process, guaranteeing robust identity anchoring.

\section{Discussion} 
\label{dis}

\noindent$\triangleright$ \textbf{\textit{Q1. Why must it be divided into three steps? What if one step goes wrong and everything follows suit?}}

The rationale for our three-stage cascaded architecture is based on explicit semantic disentanglement. Unlike end-to-end paradigms, where lighting prompts often inadvertently alter body structures, our design isolates geometry, appearance, and photometry. This ensures that the generated pose is physically aligned with the text before any texture synthesis occurs. While this pipeline incurs higher time costs and theoretical risks of error propagation, our visibility-aware optimization effectively mitigates the latter. In the context of high-value e-commerce, we argue that the precision and controllability offered by this decoupled approach significantly outweigh the latency trade-offs compared to uncontrollable real-time generation.

\noindent$\triangleright$ \textbf{\textit{Q2. How does the model demonstrate generalization beyond the semi-structured training domain?}}

Our extensive inference on "in-the-wild" images confirms that the framework has acquired a generalized understanding of the 3D interaction among light, surface geometry, and semantic context, rather than merely memorizing the training distribution. Regardless of whether the prompts specify natural outdoor illumination or complex indoor lighting, the model successfully harmonizes the foreground subject with the target ambient atmosphere , demonstrating robust zero-shot capability in diverse environmental constraints.

\noindent$\triangleright$ \textbf{\textit{Q3. What is the difference between this loss and the original InfoNCE loss used in CLIP? What is the similarity function?
}}
While FashionPose adopts the exact Bidirectional Contrastive Alignment Loss used in CLIP, it applies it to a fundamentally different domain: aligning text with explicit 2D keypoints and visibility rather than dense images. Its core purpose is to act as a regularizer to resolve the "mean pose artifact" in standard pose regression. By projecting text and pose into a normalized space and computing their temperature-scaled cosine similarity, this loss forces the generator to commit to a distinct, physically accurate geometric configuration instead of outputting a blurry, averaged estimation of multiple possible poses.

\noindent$\triangleright$ \textbf{\textit{Q4. If the text-to-pose regression is deterministic rather than generative, how does it achieve the highest variance?
}}

Although the text-to-pose regression module appears deterministic, it is evaluated by generating $K=10$ samples per prompt to calculate the spatial sampling diversity of the 2D keypoints (the "Var" metric). While the authors attribute this high variance to the contrastive loss successfully mitigating the "mean pose artifact," the paper fails to explicitly explain how this deterministic mapping is broken during inference to produce distinct samples—presumably relying on the mentioned Transformer Dropout ($p=0.05$) or some form of latent space sampling—making this a notable descriptive gap in their methodology. 

\section{Limitation} \label{lim}
While we acknowledge that our multi-stage pipeline incurs higher inference latency and potential error accumulation, these are deliberate trade-offs to prioritize Identity Preservation. To address these limitations, our future work will explore knowledge distillation techniques to compress the cascaded modules into a unified, efficient framework, thereby reducing latency while maintaining control precision. Furthermore, we intend to extend FashionPose into a video generation model, enabling the dynamic visualization of synthesized fashion results across varying viewpoints with temporal consistency.

\section{Ethics Statement}\label{ethic}
In this research, we acknowledge the potential misuse of image synthesis techniques, such as ours, for generating deceptive content and spreading disinformation, a serious concern we address explicitly. However, we also note the substantial progress made in detection and prevention mechanisms in this domain. Our framework supports critical research initiatives and encourages third-party oversight, aiming to strike a balance between technological advancement and security considerations. This balanced approach promotes responsible deployment while preserving the innovation potential.

\end{appendices}
\end{document}